# Robust Object Tracking Based on Self-adaptive Search Area

Taihang Dong*[a,b], Sheng Zhong[a,b]
[a]*Science & Technology on Multispectral Information Processing Laboratory;*
[b]*School of Automation, Huazhong University of Science & Technology, Wuhan 430074, Hubei*

## ABSTRACT

Discriminative correlation filter (DCF) based trackers have recently achieved excellent performance with great computational efficiency. However, DCF based trackers suffer boundary effects, which result in unstable performance in challenging situations exhibiting fast motion. In this paper, we propose a novel method to mitigate this side-effect in DCF based trackers. We change the search area according to the prediction of target motion. When the object moves fast, broad search area could alleviate boundary effects and reserve the probability of locating the object. When the object moves slowly, narrow search area could prevent effect of useless background information and improve computational efficiency to attain real-time performance. This strategy can impressively soothe boundary effects in situations exhibiting fast motion and motion blur, and it can be used in almost all DCF based trackers. The experiments on OTB benchmark show that the proposed framework improves the performance compared with the baseline trackers.

**Keywords:** discriminative correlation filter (DCF), boundary effects, self-adaptive search area, fast motion, visual object tracking

## 1. INTRODUCTION

Visual object tracking is a classical computer vision problem with many applications, such as human-computer interaction, traffic monitoring, video surveillance [1] [2] [3]. The generic tracking task is to estimate the position or trajectory of a target in an image sequence or a video by the initial object location given in the first frame. The tracker must have the capability to generalize the object of interest appearance from the very limited set of training samples, so this problem is especially challenging when processing situations [4] exist illumination variation, occlusion, fast motion, motion blur, etc.

Nowadays, DCF based tracking approaches [5] [6] [7] [8] [9] [10] [11] have attracted many people's attention. This kind of method learns a correlation filter from a set of training samples which come from a circular sliding window operation. This circular translation operation converts the original problem in spatial domain into the frequency domain pursuing the high computational efficiency in training and detection processes.

As discussed above, the efficient computational performance is based on the operation in frequency domain. However, the transformation from spatial domain to frequency domain has the underlying assumption that the training samples have a periodic extension. This introduces unwanted boundary effects because of the periodic repetitions of training samples. The produced boundary effects from frequency operation mainly limit the DCF based trackers' performance in two important aspects. Firstly, inaccurate negative training patches reduce the discriminative power of the learned model. Secondly, the detection scores are inaccurate around the region border while the detection scores are heavily influenced by the inaccuracy of the detection samples. Inaccurate detection scores lead to a limited target search region at the detection step.

To solve this problem, Galoogahi et al. learn model filter with circular correlation from big patches to reduce the proportion of examples in a correlation filter that are affected by boundary effects [12] [13] [14]. Danelljan et al. modify the learning objective function to penalize non-zero filter values outside the object bounding box in padding window [15]. Due to the modification of objective function, the convenience of closed-form solution disappears, so the filter should be solved by iteration causing extra obvious computational burden.

Boundary effects are inevitable when we compute score map in frequency domain. The existing algorithms focusing on handling boundary effects either only consider the learning process unilaterally or aggravate the computational load. In this paper, we propose a framework to handle the problem without damaging the closed-form solution of correlation filter. The proposed approach is achieved by changing the filter search area adaptively. In other words, we change the filter size adaptively to attain the self-adapting tradeoff between the tracking performance and computational efficiency. This framework is generic and can be incorporated into any DCF tracker without this module.

*taih.dong@gmail.com

We propose an efficient method for alleviating the boundary effects in challenging image sequences. The main contributions of this work include the following aspects:

1) A novel strategy is proposed to alleviate the boundary effects. Changing search area adaptively to fit different object movement states could circumvent the boundary effects.
2) To avoid changing search area too frequently and to keep stability, we analyze difference among various threshold value distributions and choose hysteresis function pattern to determine whether to modify the search area.
3) We compare our approach with baseline trackers in literature. Our framework could not only track the object in real time, but achieve the remarkable performance as well.

The rest of this paper is organized as follows: Section 2 presents the proposed framework. Experiment results for baseline trackers with and without our framework are demonstrated and analyzed in Section 3. Finally, Section 4 draws the conclusions.

## 2. THE PROPOSED ALGORITHM

### 2.1. Standard DCF Tracker

Before the detailed discussion of our proposed framework and for completeness, we first review the details of conventional DCF based tracker. The key contribution leading to the success of DCF based trackers is their sampling method. DCF based trackers allow for dense sampling around the target at very low computational cost through circular shifts.

The goal of discriminative learning in DCF based trackers is to learn a discriminative correlation filter that can be applied to the region of interest in consecutive frames to infer the position of the target (i.e. the location of maximum filter response). The optimal correlation filter $h^l$ is obtained by minimizing the sum of squared errors:

$$\varepsilon = \sum_{l=1}^{d} \left\| h^l * f^l - g \right\|^2 + \lambda \sum_{l=1}^{d} \left\| h^l \right\|^2 . \quad (1)$$

Here, $f^l$ is expressed as image patch feature with feature dimension index $l \in \{1, \ldots d\}$. The desired correlation output $g$ is constructed as a Gaussian function with its peak located at the target center in $f$. The $*$ represents the convolution operation and the parameter $\lambda \geq 0$ controls the impact of the regularization term. The each layer of filter $h^l$ corresponds to each dimension of image features for multidimensional features so the filter could be solved with features on each dimension.

Meanwhile, we could use various historic patches to train filter. Then the optimal filter can be computed by minimizing the output error over all training patches [12] [16]. For convenience we only consider one training sample here.

For computational efficiency, we solve (1) in frequency domain. The solution to (1) in frequency domain is:

$$H^l = \frac{\overline{G} F^l}{\sum_{k=1}^{d} \overline{F^k} F^k + \lambda} . \quad (2)$$

Here, the upper case $H$, $F$ and $G$ indicate the filter, feature and ideal correlation output in frequency domain respectively. The bar $\overline{\bullet}$ denotes the complex conjugation.

Bolme et al. have analyzed that the regularization parameter $\lambda$ alleviates the problem of zero-frequency components in the spectrum of training feature $f$ which would lead to error from division by zero [5].

To obtain a stable and robust approximation, the tracker updates the numerator $A_t^l$ and denominator $B_t^l$ of the correlation filter $H_t^l$ in (2) separately:

$$A_t^l = (1-\eta) A_{t-1}^l + \eta \overline{G_t} F_t^l \quad (3)$$

$$B_t = (1-\eta) B_{t-1}^l + \eta \sum_{k=1}^{d} \overline{F_t^k} F_t^k . \quad (4)$$

Here, $\eta$ is a learning rate parameter to control the filter updating rate. The subscript $t$ stands for the frame index.

In detection phase, given an image patch $z$, the correlation score map $y$ at a rectangular region $z$ of a feature map are computed using (5):

$$y = \mathcal{F}^{-1}\left\{ \frac{\sum \overline{A^l} Z^l}{B+\lambda} \right\}. \tag{5}$$

Here, $Z$ denotes rectangular region in frequency domain, and $\mathcal{F}^{-1}$ stands for inverse Fourier transformation operation. New target position is then located by searching the maximum of the score map $y$.

### 2.2. Proposed Framework

Boundary effects not only impact the tracker ability to detect the object position in broad area in detection process but affect the discriminating capability of model filter in training step. So we propose to add a component of changing search area in DCF based algorithms. We estimate the object movement situation and quantitative object states. Then the filter's size would be expanded when the object moves fast, which allows the filter to localize the target in a broader area mitigating the influence of boundary artifacts. The filter's size would be shrunk to increase computational efficiency while object moves slowly.

Different from the proposed way to mitigate boundary effects that tracker punishes the filter around image patch border [15] or pads zero elements around the search area [13] while detecting object, we alter the filter size adaptively to achieve this goal. We need to estimate object movement state and choose the appropriate occasion to update the size of search area. The criterion estimating object movement state is elaborated in section 2.2.1. We list the different threshold settings and analyze the difference among these threshold settings in section 2.2.2. The methods to modify filter are introduced as follows in section 2.2.3. In section 2.2.4, the pipeline of the whole algorithm is presented.

#### 2.2.1. Criterion to estimate object movement state

We expect to find a fair standard to assess the object movement state determining whether object moves slowly or not. In order to define a criterion which represents the object movement rate in images, we consider using fitting curve from last few frames to estimate the object movement state in the current frame.

Object velocity is used to curve fitting for estimating the object movement state. We use the basic polynomial curve to approximate the real object movement situation. Polynomial set is a set of complete orthogonal basis in normed space, so polynomial curve under enough orders or degrees of freedom could represent any continuous function in Hilbert space. The polynomial function fitting formula is:

$$f(idx) = \sum\nolimits_{i=0}^{vl} a_i idx^i \,. \tag{6}$$

Here, $a$ denotes polynomial coefficient and $idx$ is an independent variable with range in $\{1, 2, \ldots, n\}$ fitting function. The sign $vl$ stands for the order of the fitting curve. Superscript $i$ of $idx$ represents power of the independent variable and subscript $i$ of $a$ is an index of the polynomial coefficient of fitting curve. After the curve fitting, $f(n+1)$ is computed and then object velocity could be estimated to equal to $f(n+1)$ in the current frame.

For convenience and practicability, we fit the object velocity in x-axis and y-axis respectively:

$$f_x(idx) = \sum\nolimits_{i=0}^{vl} b_i idx^i \tag{7}$$

$$f_y(idx) = \sum\nolimits_{i=0}^{vl} c_i idx^i \,. \tag{8}$$

Here, $f_x(idx)$ and $f_y(idx)$ is x-axis component and y-axis component of target velocity, while $b$ and $c$ are coefficients of fitting functions for x-axis and y-axis respectively. We fit the polynomial function and substitute the variable into the equation to get the estimation in x-axis component and y-axis component of object velocity. Then we compute the measure using formula:

$$\zeta = \sqrt{\left(\frac{v_x^t}{sz_w}\right)^2 + \left(\frac{v_y^t}{sz_h}\right)^2} \,. \tag{9}$$

Where $sz_w$ and $sz_h$ represent the width and height of initial target size respectively, $\zeta$ is the measure reflecting object movement state. This indicator could be helpful to determine when to change the filter size.

### 2.2.2. Threshold setting

As long as the model filter is changed, it will degenerate. The more frequently the filter size is altered, the more information of object appearance the model filter loses. In order to improve the tracker robustness, the filter needs to be modified carefully. We consider different threshold settings to expand the ranges that model filters of different sizes cover in measure aforementioned.

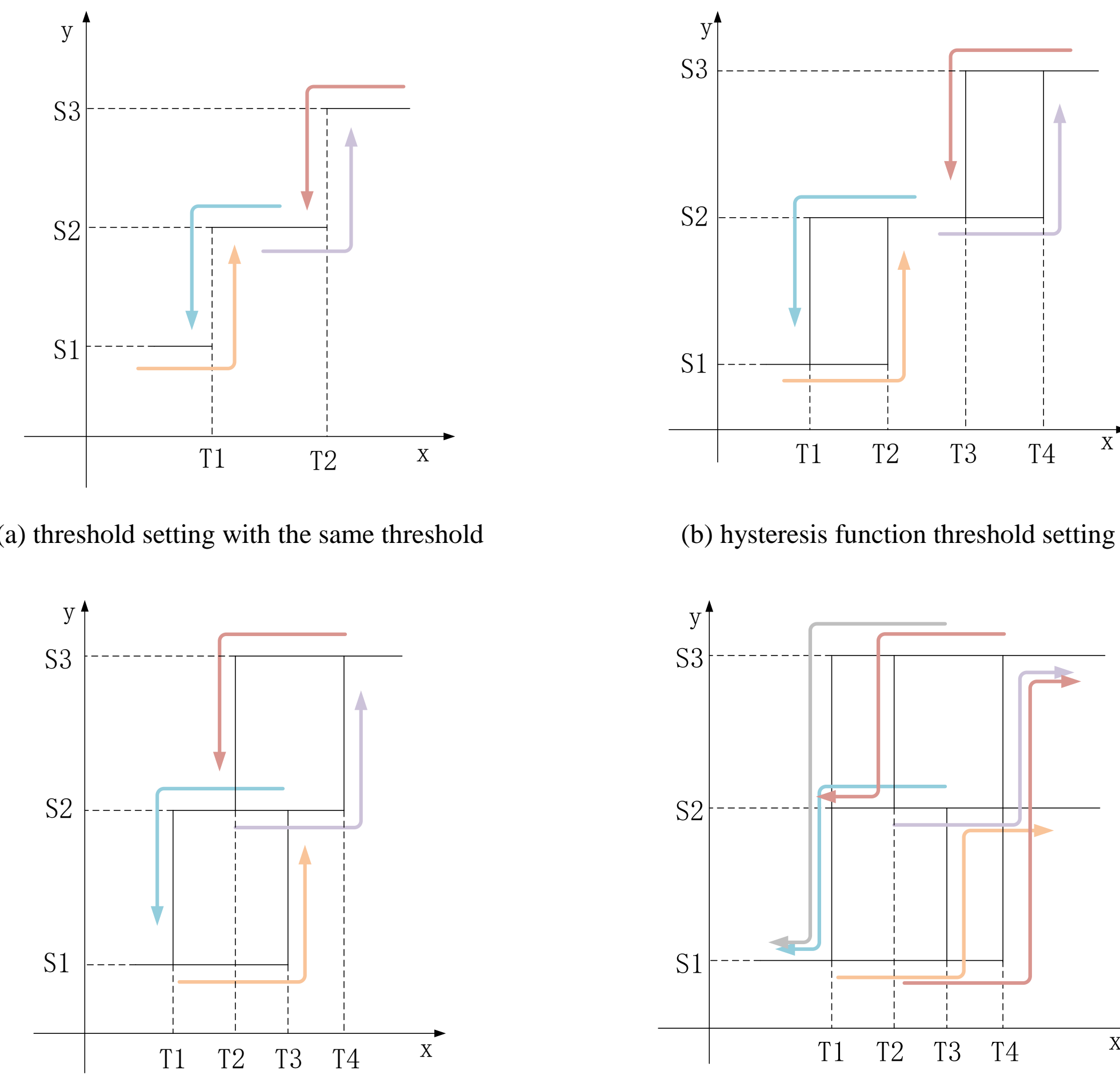

(a) threshold setting with the same threshold

(b) hysteresis function threshold setting

(c) hysteresis function threshold setting with thresholds entangled

(d) six cases of changing filter sizes

Fig. 1 The diagram about relationship between thresholds and filter sizes in different threshold settings.

First and easiest threshold setting uses the same threshold to separate the different filter sizes. There are two different thresholds to distinguish three kinds of filter size. Filter size or search area is set to S1 while criterion value is less than threshold T1. When criterion value is calculated between thresholds T1 and T2, filter size is set to S2. In the same way, criterion value greater than threshold T2 corresponds to filter size S3. The relationship between thresholds and filter sizes is illustrated in Fig. 1(a).

Threshold setting with the same threshold has the sensitive response to the criterion threshold, and this may lead to quick degeneration of model filter. So we consider the other strategy, hysteresis function, which could change search area more discreetly. When criterion value grows from small to big, the contributing thresholds are T2 and T4. When criterion value changes in the opposite direction, from big to small, the thresholds T1 and T3 play a part in the choosing of search area. Fig. 1(b) illustrates the threshold setting, and the arrow pointing represents the direction of filter size changing. Hysteresis function strategy could reduce the times of changing search area and then remarkably improve the system robustness.

Hysteresis function strategy could cautiously determine to change model filter size compared to the first threshold setting. The reason is that hysteresis function could extend the ranges that model filters of different sizes cover. So we consider further extending these ranges that model filters of different sizes cover. The threshold in hysteresis function is adjusted for this goal. The only difference between the new threshold setting and hysteresis function aforementioned is that thresholds in threshold setting get entangled. The drawing of thresholds blending together is shown in Fig. 1(c). When criterion value reflecting object movement state changes bigger, the thresholds T3 and T4 decide whether to alter the search area. When criterion value changes smaller, thresholds T1 and T2 have the right to impact the filter size.

Because the criterion value between the continuous frames is discrete and there are three kinds of filter sizes, there are six cases of changing filter sizes totally. In Fig. 1(d), we can see these six different kinds of filter size transformation marked by arrow pointing. The graph between T1 and T3 in the x-axis and between S1 and S2 in the y-axis represents the filter size conversion between S1 and S2. The graph between T2 and T4 in the x-axis and between S2 and S3 in the y-axis represents the filter size conversion between S2 and S3. Lastly, the graph between T1 and T4 in the x-axis and between S1 and S3 in the y-axis represents the direct filter size conversion between S1 and S3.

We can tell the hysteresis function with thresholds entangled could carefully change the filter size and possesses the powerful capability to improve the algorithm robustness. So hysteresis function with thresholds entangled is finally chosen to be used to determine whether to change the filter size.

#### 2.2.3. Measure to change searching area

Our aim is to update the filter size adaptively. Because the object's size in the search area can't be impacted by updating filter size, changing filter size is essentially equal to changing padding window size. One of methods to change the size of the search area is operating directly in spatial domain, and there is the other operation to change filter size in frequency domain.

For clarity, padding zero around the model filter or cropping model in the center of filter directly in spatial domain is called spatial method. Spatial method is similar to the way in [13]. The advantage of ours is that the boundary effects are alleviated in both training and detection processes. Meanwhile, we do not destruct the closed-form solution of the original algorithm to keep real-time performance. Padding zero or cropping the filter inherits the information from the last filter, which means the filter is trained using historic image patches indirectly. That guarantees the algorithm reliability.

In the same way, modifying the filter's size in frequency domain directly is called the frequency method. We interpolate the model filter using trigonometric polynomials [17] in frequency domain, which has the same result as the spatial method because of the dual character of Fourier Transformation. The tool in frequency domain is reasonable if we regard the frequency spectrum of the object as the model filter. Implicit padding zeros in spatial domain does not draw noise into the object frequency spectrum and does not discard the object's historical information. We choose to use frequency method to change filter size.

It is worth noting that the precondition of altering filter size is rational and reliable object location. The reason is to avoid the abrupt degeneration of model filter. We use the hysteresis function with thresholds entangled aforementioned to carefully change the filter size avoiding the obvious degeneration of model filter. When the object moves too fast or the object is a blur, using the adaptive search area can not only reserve the possibility to seek the right target position and alleviate the boundary effects, but also keep the computational efficiency.

#### 2.2.4. Tracking pipeline

The whole DCF based algorithm with our framework is described as follows. **Localize.** Localization estimation is same as the corresponding part in conventional DCF based trackers. **Update.** This module includes two main parts. One is updating padding size, and the other is updating filter under the updated padding size computed in advance. Updating padding size needs us to compute the object movement criterion value and use threshold setting with thresholds entangled to determine the most suitable filter size. The updating filter part modifies the original filter using updated filter size first, then updates the filter using current object appearance. The tracking iteration is summarized in Algorithm 1.

**Algorithm 1:** self-adaptive search area tracker

**Require:**

Object position $P_{t-1}$, filter $h_{t-1}$, filter size $FS_{t-1}$ in the previous frame.

**Ensure:**

Position $P_t$, filter size $FS_t$ and updated models $h_t$ in the current frame.

**Localization estimation:**

1. New target location $P_t$: position of the maximum in correlation between $h_{t-1}$ and image patch features $f$ extracted on position $P_{t-1}$ (Section 2.1).

**Update:**

Update padding size

2. Compute the criterion value $\zeta$ (Section 2.2.1).
3. Substitute the variable using $\zeta$ in hysteresis function with thresholds entangled, compute the suitable search area $FS_t$ (Section 2.2.2).

Update filter

4. Update original filter to new size $FS_t$ (Section 2.2.3).
5. Update filter $h_t$ using current object appearance (Formula (3) and Formula (4)).

We use the image sequence "soccer" in OTB to illustrate the pipeline of DCF based tracker using self-adaptive search area component in Fig. 2. The first line in Fig. 2 shows the 51st, 52nd, 53rd, 54th frame of the video "soccer" from left to right. The second line includes the response maps and model filters. The object position in the first line corresponds to the response map. The other part, the circles, is the self-adaptive search area component.

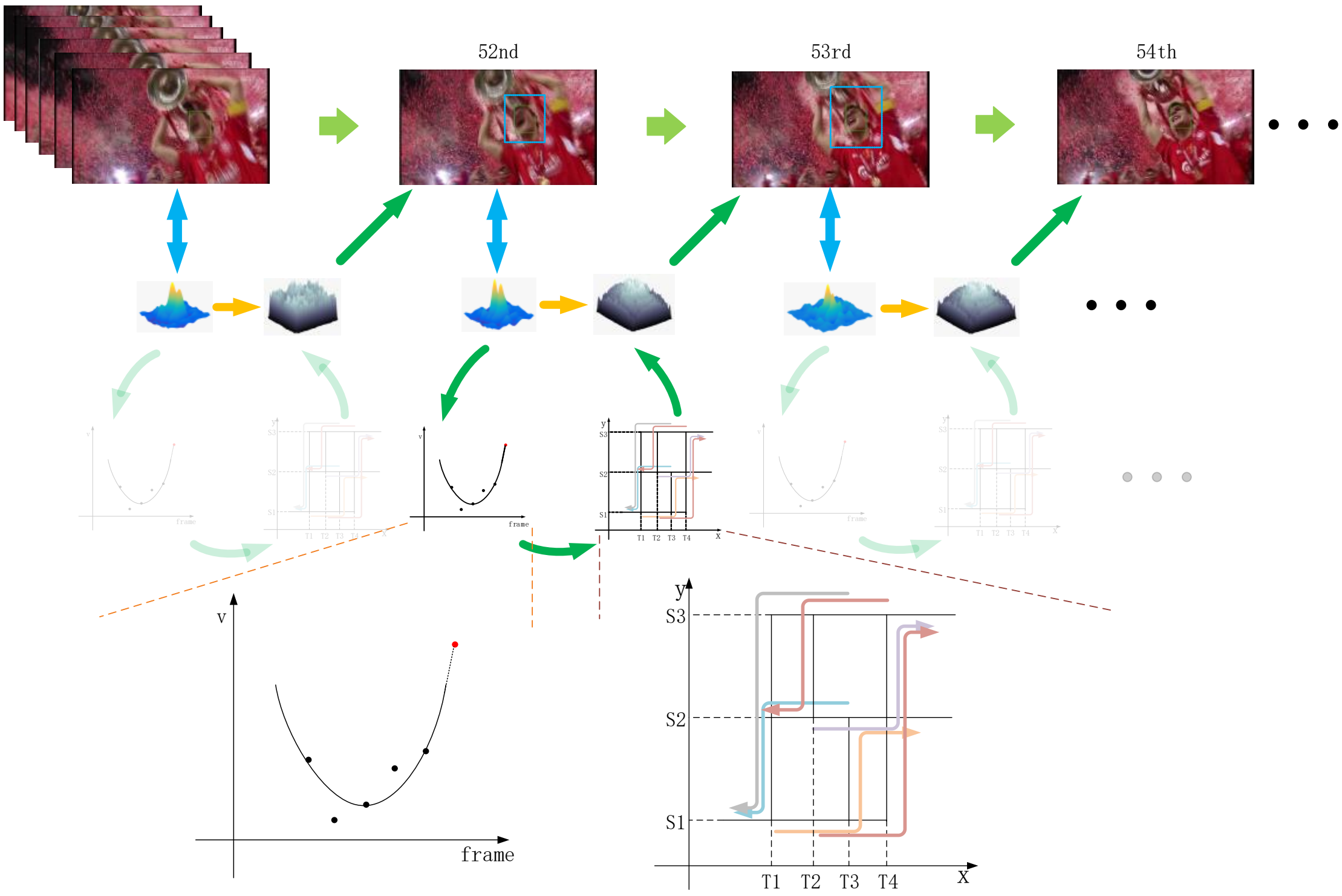

Fig. 2 The pipeline of the whole algorithm.

After updating model filter in the 51st frame, same as the conventional correlation filter tracking, the object is located in the 52nd frame. When it comes to a new frame, the object movement state needs to be estimated. Based on the 52nd frame object position, we would estimate object motion speed in the next frame. Parabola or (second order) fitting curve is used to fit the object motion speed. Object motion in last five frames is considered. After curve fitting, the object motion could be estimated easily in next frame. We substitute the next frame variable into the fitting curve to calculate the object motion speed. On the bottom of the Fig. 2, we could see the fitting curve function graph and the red point represents the object motion estimation in next frame. Then, the criterion value mentioned in section 2.2.1 is computed based on the object movement statement estimation in the next frame. After substituting the criterion value in the hysteresis function the optimal filter size is calculated. Finally, we can see the filter's size changes after this module. The filter size changes larger to track object steadily and safely. Afterwards, the model filter is updated to determine the object location in the 53rd frame like the conventional DCF based trackers do. The response map changes larger correspondingly in the 53rd frame. The same element repeats every frame and the self-adaptive search area component could find the most suitable filter's size to keep the high tracking performance and high computational efficiency.

## 3. EXPERIMENTS AND DISCUSSION

Extensive evaluations are performed on OTB benchmark datasets [4] [20] to validate the effectiveness of our framework. Section 3.1 presents three popular and diverse DCF trackers as the baseline trackers in our experiments. In section 3.2, we describe the used benchmark and details about parameters used in our experiments. We benchmark the trackers with our framework against their baseline versions in section 3.3. A comparison in attributes between our methods and baseline trackers is given in section 3.4.

### 3.1. Baseline Trackers

In order to represent the realm of DCF trackers, a wide variety of recent DCF trackers are selected as baselines. We only select trackers that follow the standard DCF formulation (1). Table 1 summarizes these DCF trackers. We apply our framework to these selected baselines and call them DCF_SASA, DSST_SASA, SAMF_SASA.

Table 1 Baseline DCF trackers to be added to our framework.

| Tracker | Features | Scale | Published |
|---|---|---|---|
| DCF | HOG[18] | No | 2015 (TPAMI) |
| DSST | HOG[18] | Yes | 2016 (TPAMI) |
| SAMF | HOG[18], CN[19] | Yes | 2014 (ECCV-W) |

### 3.2. Experimental Setup

**Evaluation Methodology.** All trackers are quantitatively evaluated on the Online Tracking Benchmark (OTB100) dataset using the evaluation protocol described in [4] [20]. OTB100 dataset extends OTB-2013 and includes 100 challenging image sequences. The tracking outcomes on the OTB dataset are reported using two standard metrics, precision and success rate, over all 100 videos. Precision measures the center error between tracker bounding box and ground truth bounding box. In the precision plot, the maximum allowed center error in pixel distance is varied along the x-axis and the percentage of correctly predicted bounding boxes per threshold is plotted on the y-axis. The common threshold of 20 pixels [4] [7] [20] is used for ranking trackers. Success rate is measured as the intersection over union (IoU) of the tracker bounding box and the ground truth. In the success plot, the required overlap is varied along the x-axis and the percentage of correctly predicted bounding boxes per threshold is plotted on the y-axis. Trackers are ranked by the area-under-the-curve (AUC) [4] [20]. We illustrate the area-under-the-curve (AUC) score and focus on success plots for conclusions and more detailed analysis rather than precision plots for reference only, since success plots are more indicative of actual tracking performance. All trackers are implemented in MATLAB, and all experiments are performed on the same computer (Intel Core i7 3.4 GHz CPU with 8 GB RAM).

**Parameter Settings.** For tracking approaches and strategies presented in section 2, the same parameter setting is used for all experiments and videos. All baseline trackers run with the fixed parameters provided by the authors. We run our self-adaptive search area trackers with the same parameters for the fair comparison. The padding parameter setting is arranged in Table 2. Note that the parameters Padding S1 are set to equal to standard padding parameters provided by the corresponding authors to achieve a reasonable and fair comparison.

Table 2 padding parameter setting.

| Parameter | DCF_SASA | DSST_SASA | SAMF_SASA |
|---|---|---|---|
| Padding S1 | 1.5 | 1.0 | 1.5 |
| Padding S2 | 2.0 | 1.8 | 2.0 |
| Padding S3 | 2.5 | 2.6 | 2.5 |

The thresholds setting corresponding to the padding parameters are summarized in Table 3. It is worth noting that, in practice, the strategy using curve fitting of velocity reduces the filter size if the criterion value is under the threshold in continuous 10 frames, and augments the search area once criterion value surpasses the fixed threshold. The sensitiveness to large search area conforms to common sense that the object moves slowly only if the measure from object movement prediction is under the threshold steadily. The image sequence is categorized to fast motion class once the object moves fast slightly causing criterion value greater than the threshold. We use object velocity of last 5 frames to fit a quadratic curve for inferring object velocity in the current frame.

Table 3 threshold setting with thresholds entangled.

| Threshold | DCF_SASA | DSST_SASA | SAMF_SASA |
|---|---|---|---|
| T1 | 0.1 | 0.1 | 0.1 |
| T2 | 0.2 | 0.2 | 0.2 |
| T3 | 0.5 | 0.6 | 0.5 |
| T4 | 1.5 | 0.9 | 1.3 |

### 3.3. Baseline Comparison

Fig. 3 show the results of all baseline trackers and their self-adaptive search area counterparts on OTB-100.

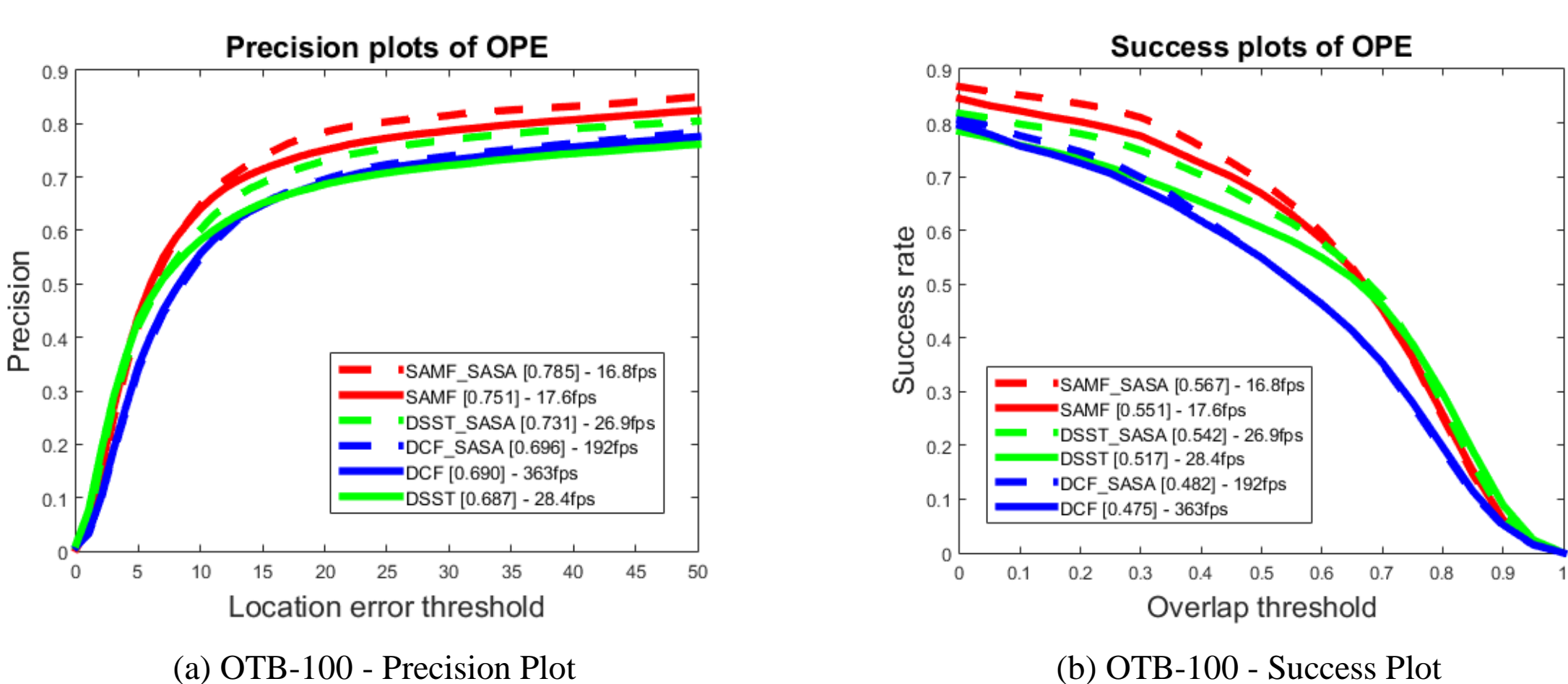

(a) OTB-100 - Precision Plot (b) OTB-100 - Success Plot

Fig. 3 Average overall performance on OTB-100. The numbers in square brackets from the left and right legends indicate the precision scores and average AUC scores respectively. The "SASA" is the abbreviation for "self-adaptive search area".

We could see that all the SASA trackers have the better performance than the baseline trackers. The absolute improvement in precision and success rate could achieve {4.4 %, 2.5%} for DSST tracker. The "fps" indicator represents the algorithm frame rate. We could tell the performance gain is achieved at a much lower computation cost, which effects little to frame rate of baseline trackers. Self-adaptive search area strategy could find the balance between the potentiality to find the object and the computational efficiency. Broad search area exploits the background information to train filter and reserves the probability to locate the object, while narrow search area improves algorithm efficiency and prevents the background clutter. Adaptively updating filter size could assemble these advantages into baseline trackers.

### 3.4. Attribute-based Comparison

We perform an attribute-based analysis of our approach on the OTB100 dataset. All the 100 videos from this benchmark are annotated with 11 different attributes, which are illumination variation, scale variation, occlusion, deformation, motion blur, fast motion, in-plane rotation, out-of-plane rotation, out-of-view, background clutter and low resolution. Here, success plots of four different attributes are illustrated in Fig. 4 which are fast motion, motion blur, scale variation, background clutter.

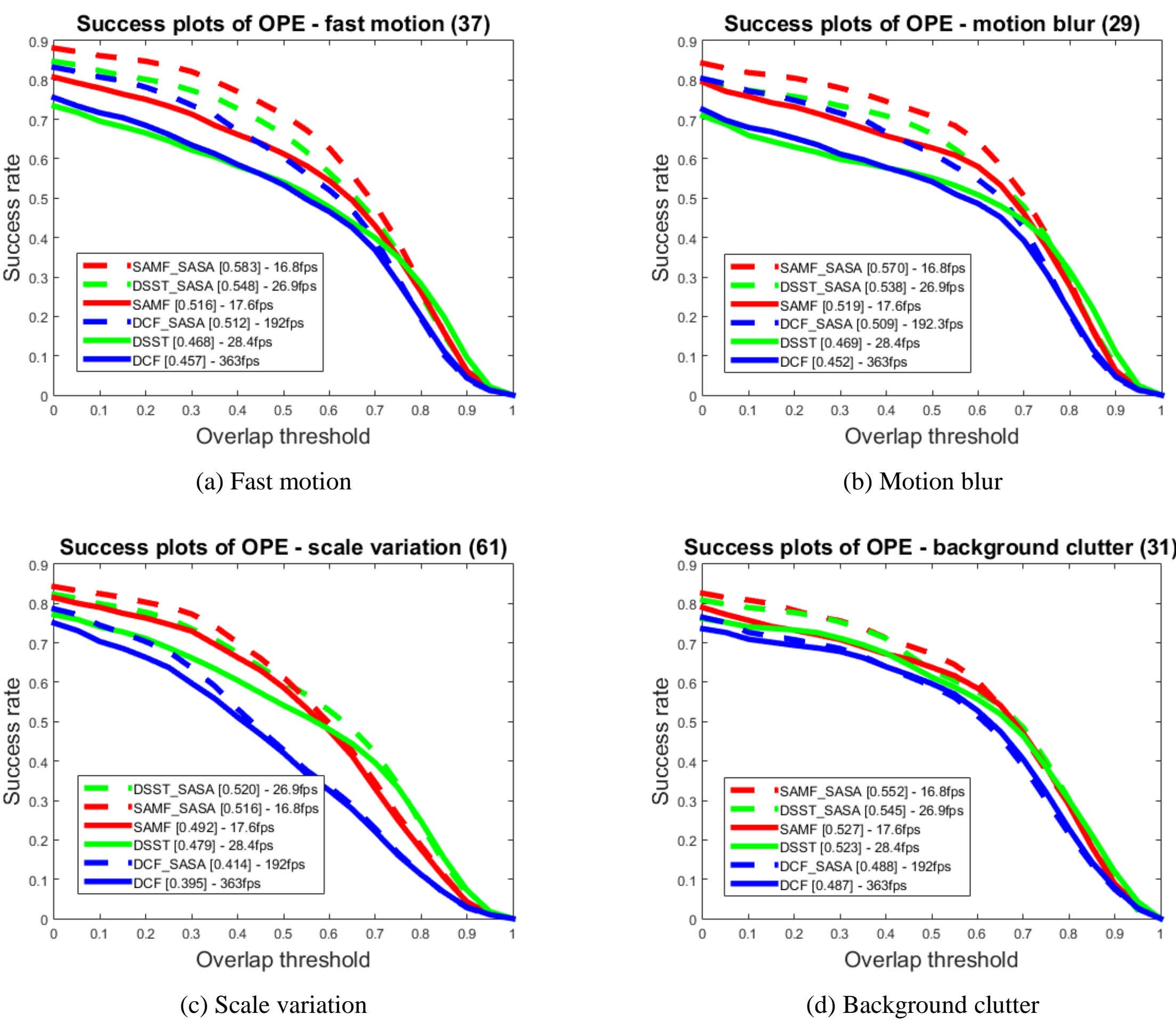

Fig. 4 Average performance on OTB-100 for 4 attributes. Each plot title shows the number of videos associated with the respective attribute.

While our framework improves tracking performance in most scenarios, what is strikingly noticeable is that there are certain categories that benefit more than others. In the case of fast motion (Fig. 4a), SAMF_SASA, DSST_SASA and DCF_SASA achieve an AUC score of 58.3%, 54.8% and 51.2% respectively, while providing gains of 6.7%, 8% and 5.5% compared to the respective baseline trackers. In the other three attributes-motion blur (Fig. 4b), scale variation (Fig. 4c) and background clutter (Fig. 4d), our framework is very beneficial similarly. For the better performance in scale variation and background clutter, this is largely due to the fact that capability of adjusting search region allows dealing with background information well. Our framework demonstrates significantly superior performance compared to baseline trackers in these scenarios.

## 4. CONCLUSION

Tracker using the big search area could find the more accurate location avoiding boundary effects, but it also causes some problems: increasing the computational cost and the likelihood of being disturbed by the noise or background around the object. So we propose a novel framework to change the size of the detection search area adaptively. Object movement state is estimated first and then whether to change filter search area is decided based on the criterion value from the estimated object movement state. When the object is estimated to move slowly, using small search area can speed the algorithm and improve the algorithm robustness. When the object is estimated to move fast, big search area can help the tracker to improve the capacity to traversal capacious area and reserve the potentiality to localize the object position with alleviating boundary effects. It's worth noting that our framework could be used in almost all DCF based trackers without this module to achieve higher performance.

## ACKNOWLEDGEMENTS

The authors are grateful for the support of the National Key Research and Development Program No.2016YFF0101502.